# Detecting retail products in situ using CNN without human effort labeling

Wei Yi[*], Yaoran Sun, Tao Ding, Sailing He

Center for Optical and Electromagnetic Research, Zhejiang University 310058, P.R. China

**Abstract**: CNN is a powerful tool for many computer vision tasks, achieving much better result than traditional methods. Since CNN has a very large capacity, training such a neural network often requires many data, but it is often expensive to obtain labeled images in real practice, especially for object detection, where collecting bounding box of every object in training set requires many human efforts. This is the case in detection of retail products where there can be many different categories. In this paper, we focus on applying CNN to detect 324-categories products in situ, while requiring no extra effort of labeling bounding box for any image. Our approach is based on an algorithm that extracts bounding box from in-vitro dataset and an algorithm to simulate occlusion. We have successfully shown the effectiveness and usefulness of our methods to build up a Faster RCNN detection model. Similar idea is also applicable in other scenarios.

**1. Introduction:**

Nowadays, deep learning is becoming more and more popular both in academia and industry, and it is mainly based on deep neural networks. These methods have dramatically improved the state-of-the-art on many different benchmarks. Specifically, in computer vision, CNN(convolutional neural network) architecture is mostly applied. Commonly used CNN backbones include VGGNet[1], ResNet[2], FPN[3], etc. As for deep detection models, there are fast one stage models like YOLO[23,24], SSD[25-27], and accurate two stage model like Faster RCNN[18], Light-head RCNN[28], R-FCN[29].

Detection of items from a shopping list is a problem previously explored in many fields[4-9]. One traditional method is to use RFID[30] (Radio Frequency Identification) technology to put a unique label into every product, however this kind of label is not reusable and need to label every product even though they are actually the same type of product. While computer vision technology can easily overcome these drawbacks. But to apply CNN on this retail product detection task, a large amount of training data is needed, which is also expensive in practice.

So, in this paper, we focus on the task of detecting all products in a still image both in vitro (i.e. ideal environment) and in situ (i.e. real scenes environment) using CNN network, but at the same time with no extra human effort of labeling bounding box for every image.

We construct this paper as follows: Section 2 reviews the related work on detecting products. Section 3 presents our main approaches. In section 4, experiment results of our algorithms are demonstrated and discussed. Finally, the conclusion is described in section 5.

* Corresponding author.
  E-mail address: yiwei2016@zju.edu.cn (W. Yi).

## 2. Related work

Collecting classification dataset is much easier than object detection datasets, because classification data requires no bounding box labeling, so most work was done on product recognition[10-14], i.e. given a image containing only one product, output what it is. As for detection, there might be multiply products in a single image, we need to find and locate all of them. There are traditional computer vision methods[6] and deep model based methods[4,15].

For traditional methods, color, shape and texture feature are common approaches[11]. SIFT key points and SURF descriptors[16] are also useful to represent rotation, scale and translation invariant features, and they can capture the main feature of an object that we want to detect even if there are some deformations or unforeseeable noise. After extracting those invariant features, some traditional machine learning algorithms or template matching algorithm are applied to draw a decision boundary to distinct all classes. For example, in [17], they first extract enhanced SURF descriptors from images, and then use a multiclass naive-Bayes classifier to classify products in GroZi-120[5] dataset. While in [6], they use random forest for the final classification step. One main disadvantage of those methods is that those extracted feature is based on some human domain knowledge to some extent. While for deep learning, model can learn to extract hierarchical features by itself, so it is more efficient and more useful.

For deep methods, Faster RCNN[18] is the state-of-art detection model, it is a two-stage algorithm, in stage one, region proposals are generated by a RPN network, location of every object is refined in stage two, together with classification in every refined bounding box. Since collecting detection dataset is expensive, most deep learning based detection work focus on how to train a better model using very few data(even one image per category). In [9], they perform a multi-stage training procedure, in which they first pre-train on a large class-level dataset, followed by an auxiliary multi-view dataset, which trains the network to be robust to viewpoint changes. Finally they train on the objects that they wish to recognize from just a few image. In [4], they use a non-parametric probabilistic model for initial detection to estimate a short-list of possible categories and then a deep neural network is used for refinement. This is the case when we are unable to collect enough data for training a deep detection model, but since deep model shows better performance when we have more data to train, a good way to improve our performance would be trying to extend our datasets with high quality data, but at the same time, with little extra human work to make this approach feasible in real practice. In this paper, we propose a method to solve this problem.

## 3. Approach
### 3.1 Basic dataset setup

Since collecting data in situ is much more slower, we first build up our basic dataset in vitro. We use an array of industrial cameras(9 cameras in all, 6 for top-down view, 3 for side view) to capture different point of view of every product. The setup is shown in Fig.1.

We collect both images that contains single product and multiply products. For single product images dataset, i.e. the classification dataset, 324 categories of products, 5000 images per product is collected. This dataset is used for training a classification model, we use ResNet50 as the backbone for transfer learning[31-33], we first fine tune the last few fully connected layer and then the deep convolution layers. As for multiply products images, all we need to record is only how many products in total(**N, N**>1 here, for single product images above, **N** is 1) are in the image, we do not need to know what the category labels of this **N** objects are, and we also do not need to collect the ground truth bounding box information at this stage either, those are works we will handle later on. Random combination of all 324 products is chosen and **N** is chosen in 2,3,4 and 5. In all, 6000 multiply products images are collected. Some samples are shown in Fig.1.

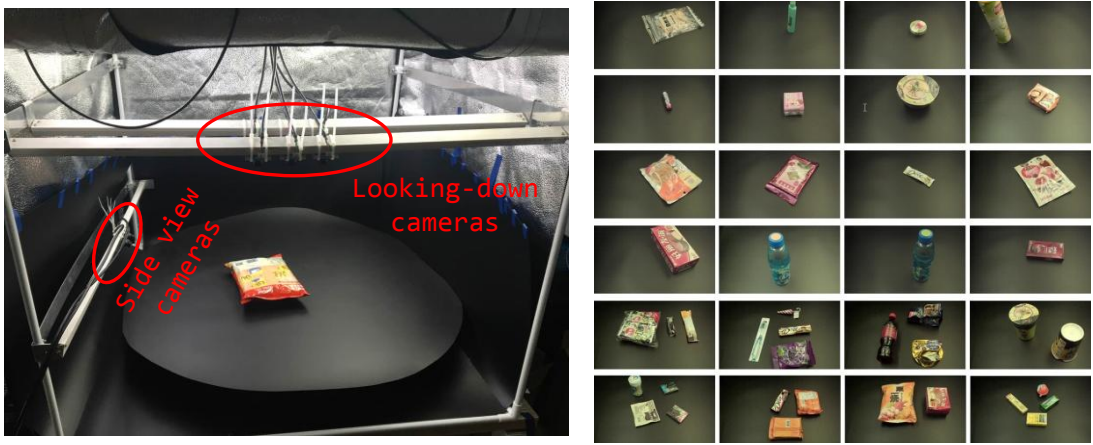

Fig.1. Hardware setup and overview of some examples from the collected datasets.

For traditional detection pipeline, The next stage is to label the dataset with bounding boxes. Note that collecting bounding boxes for those data often requires months of work. To solve this problem, we proposed **bounding box extraction algorithm**(detail is in 3.2) to do this labeling work and **occlusion simulation algorithm**(detail is in 3.3) to future augment our training set.

Since current two existing retail product benchmarks, GroZi-120[5] and GroZi-3.2K[6], have a few images per category, on average 5 and 1 images respectively, which are quite insufficient to train deep models. For future research, we will make our dataset public later on.

### 3.2 Bounding box extraction algorithm

The workflow of bounding box extraction algorithm is described in Fig2. There are two parts: the extraction part and confirming part(only for multiply products images). The purpose of extraction part is to generate bounding box candidates with high probability to be considered as ground truth, and to test whether they are really valid bounding box for this image, those candidates are passed to the confirming part.

For the extraction part, first, a bunch of bounding box proposals are generated by selective search[20], since we have the prior knowledge that we just have exactly **N** object(s) in the image, we can then merge these proposals when two proposals have an IoU(intersection over union) larger than a threshold $I_{th}$, we stop this merging process until we have got only N proposal(s) left. The use of selective search is quite different from the original pipeline of R-CNN[21,22]. In R-CNN model, we already have the labeled bounding boxes of every image, selective search is used for training and inferring, while here, we use selective search to help us get the position of bounding boxes for every image.

For multiply products images, generated bounding box by selective search can be quite complex, we use the second part to confirm the final result, specifically, we use the classification model trained by single products images to classify the object in those bounding boxes, we do not accept this bounding box unless it is classified as a valid product category and the softmax score is larger than a threshold.

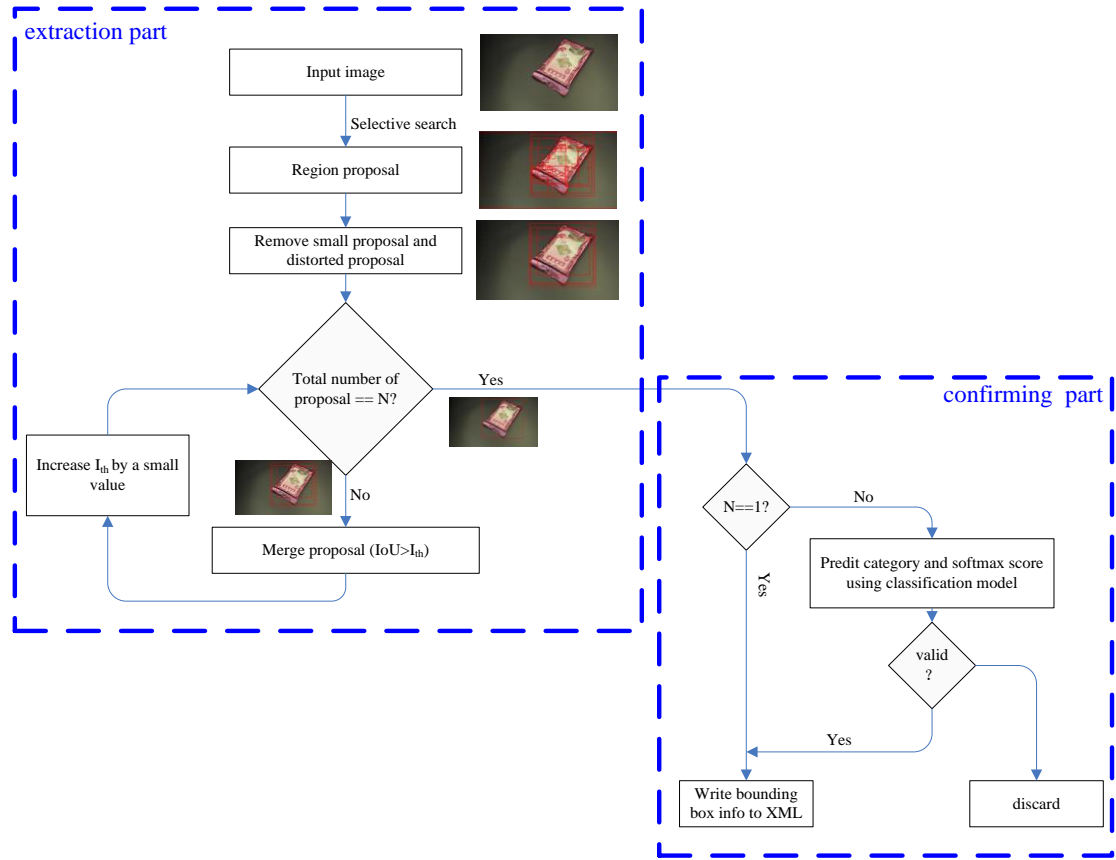

Fig.2. The workflow of bounding box extraction algorithm (best view in color)

Implementation details of extraction part is shown in the following pseudo code. Since the first step of using selective search usually results in hundreds of region proposals(denoting as **RP**), the original size of **RP** is always greater than **N**, so we can gradually merge those proposals by increasing the IoU threshold $I_{th}$ by a small value $\Delta_{th} = ((size(RP) - N)/100)$. Notice that this small value $\Delta_{th}$ is adaptive to the size of remaining proposals $size(RP)$. At first, $size(RP)$ is large, we can merge in a big step, as $size(RP)$ gradually converge to the ground truth **N**, $\Delta_{th}$ becomes small in order to avoid merging too fast to get some undesirable bounding boxes. In this way, this algorithm can be quite time efficient.

**extraction algorithm**

M is the input image

N is the exact number of objects in M

RP is a list of possible region proposals for M (each box is represented by left-up corner coordinate and width, height)

$I_{th}$ is IoU merge threshold, $A_{th}$, $D_{th}$ is threshold to filter small and distort bounding box

    RP = **Selective search**(M**)**
    **for** xmin,ymin,W(width), H(height) **in** RP**:**
        **if** W*H<$A_{th}$ or (W/H)> $D_{th}$ or (H/W)> $D_{th}$ **do**
            remove current region proposal in RP
        **end if**
    **end for**
    **while** size of RP != N **do**
        RP' = empty
        **for** P **in** RP **do** (outer **for** loop)
            **for** P' **in** RP' **do** (inner **for** loop)
                **if** IoU(P, P') > $I_{th}$ **then**
                      **go to** outer **for** loop
                **end if**
            **end for**
            add P to RP'
        **end for**
        RP = RP'
        $I_{th}$ = $I_{th}$ + (size(RP) - N)/100
    **end while**

### 3.3 Occlusion simulation algorithm

The above dataset is collected in vitro, In order to make our model more robust in situ application. We try to simulate the situation of occlusion, which is a common phenomenon in real practice. We overwrite a random region in original image either by a black block or a random patch from another product(can also be itself). Those patches are stored in a patch database, which is collected after we get the bounding box of every image. To make this simulation more efficient and more realistic, we threshold each patch to get a binary mask, we only modify values in this mask, instead of regions in the whole bounding box,

$$augmented\_image(i) = \begin{cases} patch(i) & if\ mask[i] = 1 \\ original\_img[i] & if\ mask[i] = 0 \end{cases}$$

the pipeline is showing in the following figure:

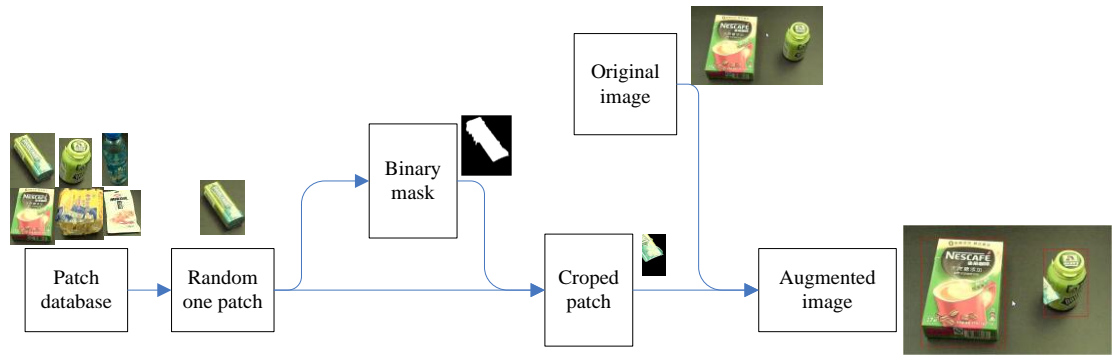

Fig.3. Occlusion simulation algorithm pipeline

## 4. Experiment

In this section, we will show some experiment results and analysis on previous methods. First, the results of extracting bounding box and occlusion simulation using algorithm presented in section 3.2 and 3.3 are shown. Then, we fine tune a faster RCNN model, the in vitro detection result is shown, after that, we test our model on in-situ environment qualitatively. To future evaluate our method of getting bounding box without human label effort, we also label several images with ground truth bounding boxes both in vitro and situ to get a quantitative result.

### 4.1 Dataset collection result

For basic classification data collected above, N is 1, so we just merge all the proposals until one proposal left, and the object category label is just the same as the original classification label, some samples are show in Fig.4. The raw detected bounding box of selective search contains many small boxes, this is due to the un-uniformity of the light intensity or the internal noise of cameras, those invalid boxes can be filtered beforehand. As we gradually increase the IoU threshold $I_{th}$ based on the total number of remaining bounding boxes, we get fewer but more accurate bounding boxes. After we are done, we can save the bounding box information to local XML files and save this patch in the bounding box to the patch database. From the results, we can see that for classification dataset where **N**=1, the bounding box extraction algorithm works quite well, the extracted bounding box is tightly close to the boundary of the product.

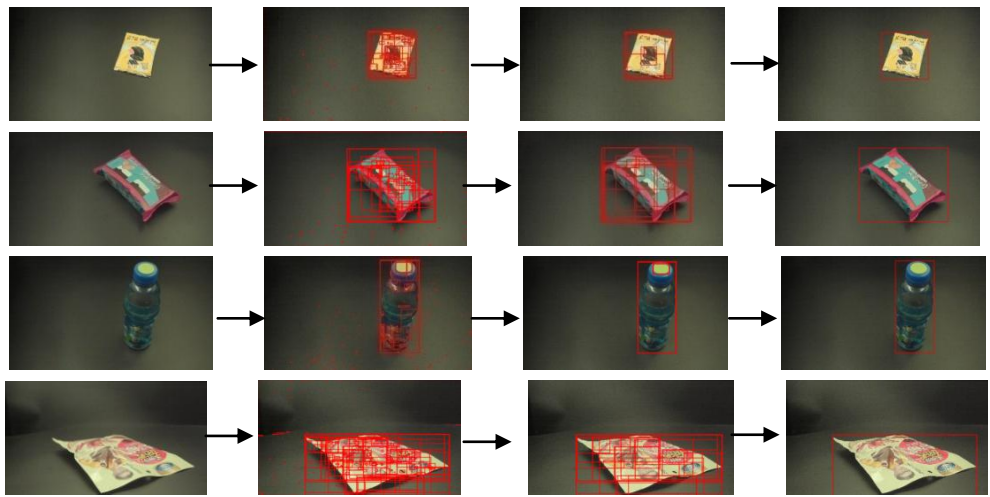

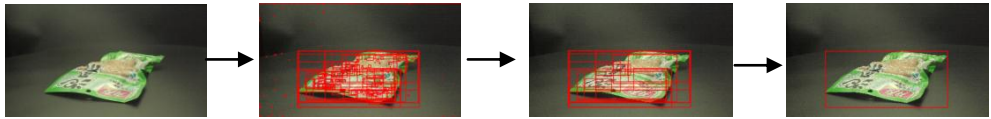

Fig.4. Examples of gradually merging bounding boxes for N=1, first 3 rows are taken from the looking-down cameras, last 2 rows are taken from side view cameras

Also, for images where **N**>1, apart from merging bounding box based on **N** as showing above, we also feed this patch of image in the bounding box to the classification model (trained by single product images) to get the label and corresponding confidence score to confirm whether the extracted box is valid or not. A threshold confidence score of 0.8 is used. Since the size of unlabeled image can be quite large, drop a small portion of those data will not influence much on the total size of collected labeled data. This confirm step will allow us to get a high-quality labeled detection dataset, which is crucial for training our model to get a good performance. Here are some successfully extracted samples.

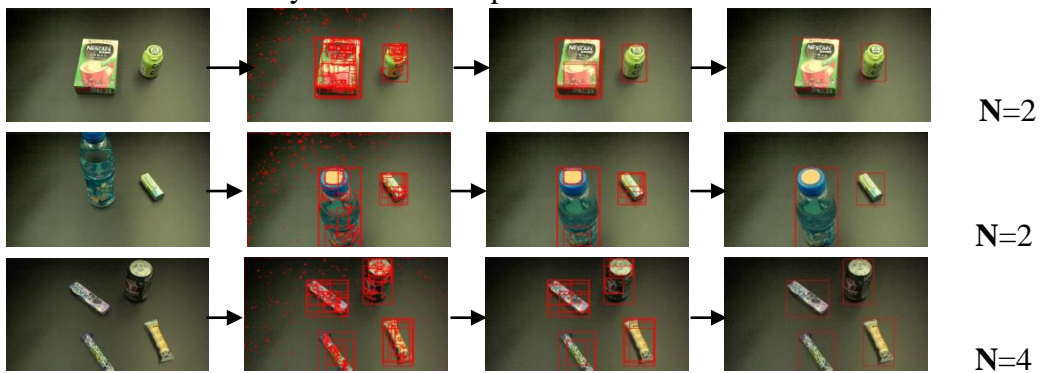

Fig.5. Examples of gradually merging bounding boxes for N>1

For multiply products images in the training set, we deliberately separate them to avoid overlapping, since overlapping will bring some trouble in finding all the correct bounding boxes for all objects in the image. If we train Faster RCNN only on the above dataset, we cannot general well to images where different products have intersections. To eliminate this problem, we use algorithm described in section 3.3 to simulate occlusion. Two different methods(black block and part of other products) and all four possible occlusion directions(left, right, up and down) are considered. Some samples are showing in Fig.6.

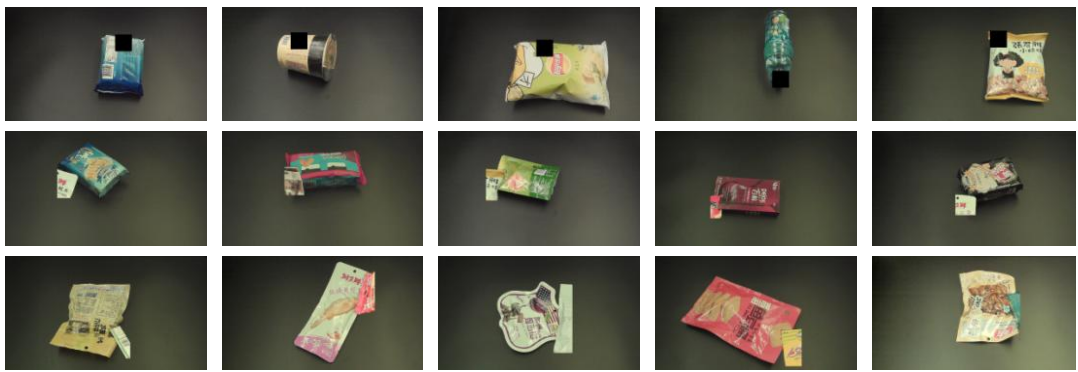

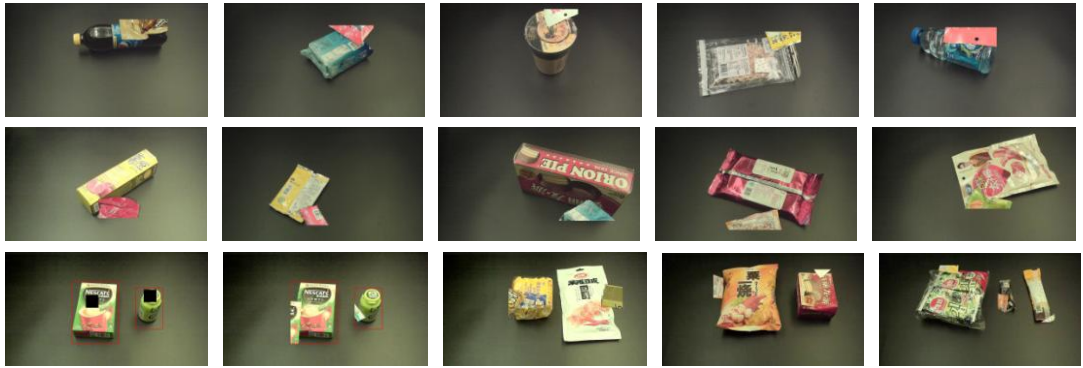

Fig.6. Examples of occlusion simulation result, 1st row use black block to fill a random region, 2nd-5th row shows how to use random patch from another product to simulate, including all 4 direction: left, right, up and down. 6th row is examples showing occlusion simulation in multi-object image

Now, we have got enough high-quality labeled training data, we can then fine tune our Faster RCNN detection model. Performance is showing as follows.

**4.2 In vitro detection result**

In this section, products are placed under the same environment as training set, we both take photos using the original industrial camera and a smart phone camera. For original industrial camera, a 640x480 pixel image is obtained, we feed the image directly into our trained detection model for inference. As for the image taken by smart phone, the size of image is 1920x1080 pixels, which is quite large for Faster RCNN to detect. To make our detection faster, we first resize it to 640x360, which preserves the aspect ratio, and then run detection on the resized image. Note that the pictures taken by this two cameras can be quite different, both due to the hardware chip used to sense the light intensity and internal software algorithm to post-process the raw image. Single object, multiply objects and objects with occlusion are all tested. Some result can be seen in Fig.7.

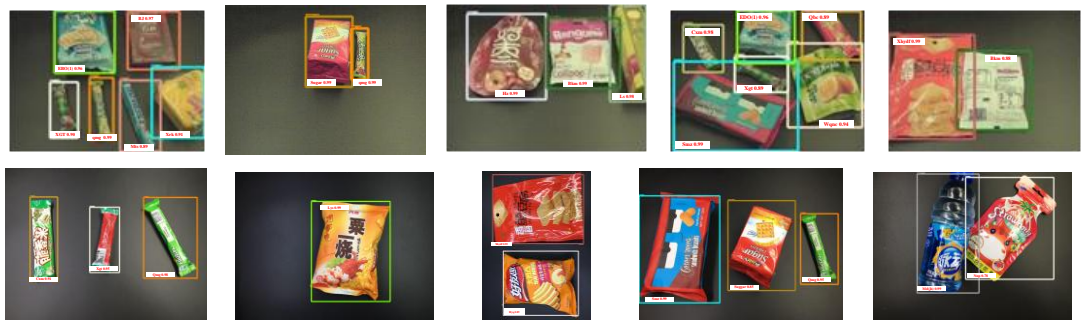

Fig.7. Examples of product detection results. The first row is taken by industrial camera, the second row is taken by a smart phone. Each detected product is drawn by a bounding box, the corresponding category label and a softmax score in [0:1], score under 0.7 is suppressed.

From the detection result, we can see that with enough training data, our trained model can be very general to different cameras, showing that our model has learned something useful from our dataset, which indicates that our extraction and simulation algorithm are quite efficient. We will future explore how our model can transfer to other different in-situ environment in section 4.3.

Even there are some overlapping between products, our trained model is also able to predict correctly what are in those images either using the original environment or use smart phone, because we have adopt algorithm presented in 3.3 to simulate real occlusion situations.

To quantitatively test our algorithm, we have also labeled 400 images both taken by original industrial camera and smart phone camera. An mAP of 0.86 and 0.83 is got respectively. Other metrics including $AP_{max}$(maximum AP over all categories), $AP_{min}$ (minimum AP over all categories) and average recall are shown in table 1. Those results show that our generated dataset has a high quality and is sufficient to fine tune a Faster RCNN model. Since all bounding box collecting process requires no human labeling efforts, it proves the usefulness and effectiveness of our algorithms.

**table.1.** quantitative results of detection model trained on generated dataset

|  | in vitro original camera | in vitro smart phone | in situ conveyor | in situ shelf |
|---|---|---|---|---|
| **mAP** | 0.86 | 0.83 | 0.84 | 0.79 |
| **$AP_{max}$** | 0.96 | 0.94 | 0.96 | 0.88 |
| **$AP_{min}$** | 0.79 | 0.77 | 0.77 | 0.71 |
| **recall** | 0.95 | 0.94 | 0.95 | 0.90 |

### 4.3 In situ detection result

In our previous settings, bounding box extraction algorithm and occlusion simulation algorithm are applied in vitro images, while in real supermarket, we sometimes also want to detect products in situ. To future exam our methods, we consider two common cases: products on conveyor and products on shelf. One point to emphasis is that our detection model is trained only on the in-vitro dataset, while test set in this section is all collected in situ.

For the case of conveyor, products are transported into the detection region of a top-down side view camera, note that the color of conveyor is green and has a limited perspective. And environment of shelf is also quite complicated, including the light intensity variation, different background and the large density of products. These cases are quite different from in-vitro environment settings.

Thanks to CNNs' great capability of generalizing to other tasks, and equipped with our large training set collected without human efforts, we can still handle this problem. Using the model described above, we get an mAP of 0.84 on conveyor and 0.79 on shelf. Some in-site examples are shown in Fig.7. And detail metrics result are shown in table.1.

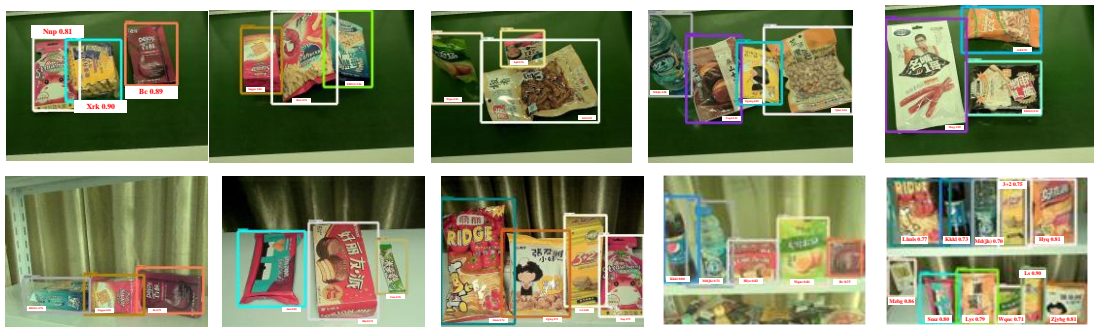

Fig.8. Detection result for in-site samples

We can see that even in a totally different environment, our model can also perform quite well. This result proves that our model trained on our generated dataset is transferable to various complicated environments, future indicates the high quality of our generated dataset and correctness of our algorithms used to extract bounding box and simulate occlusion.

## 5. Conclusion & future work

In this work, we have successfully applied CNN to the task of retail products detection without any human labeling efforts, we have proposed bounding box extraction algorithm and occlusion simulation algorithm to get a object detection dataset. We have trained a Faster RCNN model using this dataset, and trained model is tested both in vitro and in situ, qualitative and quantitative results are presented. We have proven the effectiveness of our proposed methods. Those similar ideas are also applicable in other scenarios where bounding box labeling is needed.

Just like Faster RCNN uses a RPN network to substitute the selective search stage, in the future, we can also apply this idea to form a unified network architecture to extract bounding boxes.